# Structured Prompt Optimization for Few-Shot Text Classification via Semantic Alignment in Latent Space

Jiasen Zheng
*Northwestern University*
Evanston, USA

Zijun Zhou
*Independent Researcher*
San Jose, USA

Huajun Zhang
*Syracuse University*
Syracuse, USA

Junjiang Lin
*University of Toronto*
Toronto, Canada

Jingyun Jia
*University of Wisconsin-Madison*
Madison, USA

Qi Wang*
*Purdue University*
West Lafayette, USA

*Abstract*-This study addresses the issues of semantic entanglement, unclear label structure, and insufficient feature representation in few-shot text classification, and proposes an optimization framework based on structured prompts to enhance semantic understanding and task adaptation under low-resource conditions. The framework first uses a pretrained language model to encode the input text and obtain basic semantic representations. It then introduces structured prompts composed of multi-dimensional semantic factors and integrates them with text features through a learnable combination mechanism, which forms task-related representations with clear boundaries in the latent space. To further strengthen the consistency between text representations and label semantics, the method constructs a structured label embedding matrix and employs a cross-space alignment mechanism to ensure stable matching between textual features and label attributes. In addition, the model applies prompt orthogonality constraints and a joint optimization objective to maintain independence across different semantic factors in the prompts, allowing the structured prompts to provide transparent and controllable guidance for classification decisions. Three types of sensitivity experiments, including learning rate sensitivity, prompt length sensitivity, and data scale sensitivity, are designed to evaluate the stability and robustness of the framework under different conditions. Experimental results show that the proposed structured prompt optimization framework effectively alleviates semantic conflicts and label ambiguity in few-shot text classification. It significantly improves performance on accuracy, precision, recall, and AUC, and demonstrates strong cross-task applicability. Overall, the method achieves stable semantic modeling under limited data and offers an efficient and interpretable optimization pathway for few-shot text classification.

*Keywords: Structured prompts; few-shot text classification; semantic alignment; pre-trained models; latent space modeling*

## I. Introduction

In recent years, text classification has become one of the fundamental tasks in natural language processing. It plays a crucial role in sentiment analysis, risk identification, public opinion monitoring, medical text understanding, and many other real-world scenarios[1]. However, real applications often lack sufficiently large and high-quality labeled datasets. Few-shot, weakly supervised, or even zero-shot settings are common, which limits the expressive power of traditional deep models. Under such constraints, developing a method that remains stable with only a small number of training samples has become a central challenge for practical artificial intelligence. On the one hand, data distributions vary across domains, contexts, and writing styles. Models trained on small datasets easily suffer from representation shift, inconsistent features, and unstable training. On the other hand, label semantics differ across tasks, and these variations make generalization even more difficult. Therefore, research on few-shot text classification is not only about algorithm design but also affects the sustainable deployment of language technologies in industry and public service[2].

The development of large-scale pretrained language models brings new possibilities to few-shot learning. Their strong semantic generation ability enables prompt learning to become an important way to enhance model generalization. Prompt structures inject task information into the input and guide the model to interpret text in a specific manner, which reduces the need for large labeled datasets. However, the latent space of pretrained models is complex and uneven. Manually designed prompts often fail to match the specific task precisely. Small differences in expressions, template forms, or keyword choices can cause significant performance fluctuations. In addition, label semantics vary across tasks and lack a consistent structure. Static prompts cannot cover the needs of multi-task, multi-domain, or multi-granularity scenarios. As a result, models often struggle with inadequate task understanding and insufficient semantic guidance under few-shot conditions. Prompt learning improves transferability in low-resource settings, but it still lacks systematic mechanisms for aligning prompts with task structure and ensuring controllable semantic guidance[3].

## II. METHODOLOGY FOUNDATION

By integrating semantic-guided adaptation mechanisms [4], the framework adopts the principle of embedding explicit semantic control into parameter-efficient components, which directly informs the construction of multi-dimensional structured prompt factors. Latent capacity regulation strategies demonstrated in [5] guide the design of controlled semantic compression, ensuring that structured prompts do not overwhelm the pretrained representation space under few-shot constraints. Efficiency-aware fine-tuning techniques [6] further support selective and stable parameter interaction, motivating the learnable combination module that merges structured prompts with base text features without destabilizing pretrained semantics.

Structured semantic enhancement approaches [7] provide methodological grounding for incorporating task-aware semantic attributes explicitly into representation learning. Fine-grained explainable representation modeling [8] supports the decision to decompose prompts into independent semantic factors rather than relying on monolithic templates. Retrieval-augmented semantic stabilization mechanisms [9] reinforce the necessity of strengthening feature-label consistency, which leads to the introduction of a structured label embedding matrix and cross-space alignment constraints in the proposed model.

Structural independence modeling [10] contributes directly to the orthogonality constraint design, where semantic prompt factors are enforced to remain disentangled in latent space. Regularized parameter-efficient optimization [11] informs the joint optimization strategy that balances prompt adaptation with representation stability. Uncertainty-aware modeling techniques [12] guide the construction of an objective that maintains prediction reliability while introducing additional structural constraints.

Cross-domain semantic alignment mechanisms [13] highlight the importance of maintaining semantic consistency across varying data distributions, supporting the robustness analysis of structured prompts under few-shot conditions. Adaptive prompt fusion strategies [14] demonstrate that dynamic integration of semantic components improves cross-task transferability, which aligns with the learnable fusion mechanism adopted in this framework. Finally, modular semantic injection techniques [15] validate the feasibility of inserting structured semantic units into pretrained models without disrupting global representation coherence, providing architectural justification for treating structured prompts as modular latent regulators.

## III. PROPOSED FRAMEWORK

In this framework, the input text x is first processed by the encoder of a pre-trained language model to obtain a basic semantic representation. Its model architecture is shown in Figure 1.

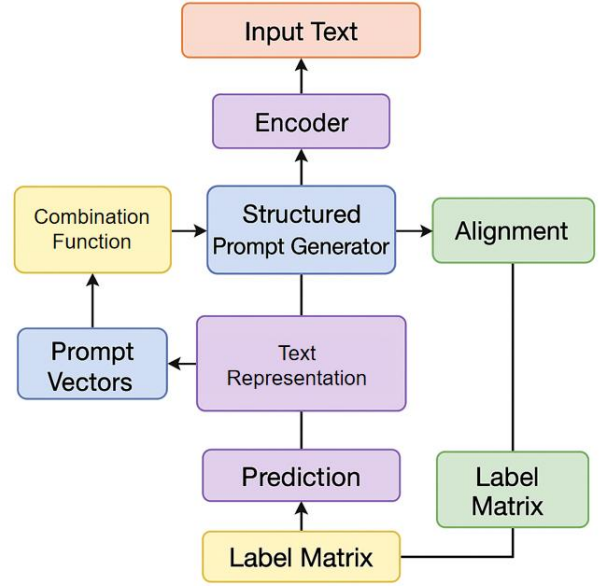

Figure 1. Overall model architecture diagram of this article

To reduce the semantic shift problem under few-shot conditions, the text representation is defined as:

$$h = f_{enc}(x) \quad (1)$$

Here, $f_{enc}$ represents the encoder of a frozen or partially trainable pre-trained model. Subsequently, to enable the model to accept semantic guidance from structured cues, a set of task-related cue vectors, $\{p_1, \ldots, p_n\}$, is introduced and fused into the input space using a learnable combination function, yielding a joint representation:

$$z = g(h, p_1, \ldots, p_n) \quad (2)$$

This fusion mechanism enables the model to simultaneously perceive text content and task structure during the encoding phase, thereby establishing more stable semantic boundaries in small-sample scenarios. To clarify the semantic structure of the labels, this method maps each label to a structured semantic vector space, constructing a label semantic matrix:

$$E = [e_1, \ldots, e_C]^T \quad (3)$$

Where C represents the number of categories, and $e_i$ is the structured representation of the i-th label. This representation includes both the semantic attributes of the label and its functional features in the task. Subsequently, a cross-space alignment function is used to strengthen the association between the text representation and the label structure, enabling the model to achieve a stable mapping in the latent space. The alignment loss is defined as:

$$L_{align} = \sum_{i=1}^{C} d(z, e_i) \quad (4)$$

Here, $d(\cdot)$ is a differentiable distance metric function used to measure the degree of fit between the text semantics and the label semantics. Through this mechanism, structured prompts can continuously calibrate the model's classification decision direction during training.

To enhance both controllability and adaptability in the structured suggestion generation module, we introduce a dynamic optimization function that iteratively updates the suggestion vector during training. Instead of treating suggestions as fixed auxiliary embeddings, the model refines them through a structured objective that accounts for task feedback and contextual alignment. The autonomous knowledge structuring framework proposed by Wang et al. [16] demonstrates that self-driven representation refinement enables intelligent agents to adapt to open-world uncertainties. Inspired by this principle, our suggestion optimization mechanism allows the suggestion vector to be updated based on semantic feedback signals, thereby improving adaptation under limited supervision. To ensure that the update process remains computationally efficient and parameter-economical, we incorporate a structured modulation strategy similar in spirit to the proactive low-rank adaptation mechanism described by Ni et al. [17]. Their Predictive-LoRA framework highlights the importance of structured and lightweight parameter adjustments for scalable large language model inference. In our method, the suggestion updates are constrained within a structured parameter subspace, balancing adaptability with training stability. Moreover, uncontrolled updates may introduce semantic inconsistency or representation drift. Gao et al. [18] show that contextual trust evaluation mechanisms improve robustness in multi-agent LLM coordination by regulating interaction dynamics. Drawing from this insight, we incorporate a contextual consistency evaluation term that governs the direction and magnitude of suggestion vector updates. This ensures that the optimization process preserves semantic reliability and task alignment. Accordingly, the structured suggestion optimization problem is formalized as:

$$p_j^* = arg \min_{p_j} L_{task}(z, y) \quad (5)$$

Where y is the target category label, and $L_{task}$ is the task-level loss, used to measure the correctness of the final classification. To ensure the independence of prompts across different semantic dimensions, a prompt orthogonality constraint is also added:

$$p_i^T, p_j = 0 \quad (6)$$

This constraint avoids semantic overlap within the prompts, thereby improving the prompts' ability to accurately guide different semantic factors and making them more suitable for efficient operation under small sample conditions.

In the final classification stage, the model matches the joint representation z with the structured label matrix E, and obtains the classification probability through a scoring function:

$$\widehat{y} = softmax(EWz) \quad (7)$$

Where W is a learnable mapping matrix used to achieve semantic alignment between the latent space and the label space. To improve the robustness of the overall framework, the method uses the task loss, alignment loss, and cue regularization term together to form the overall optimization objective:

$$L = L_{mask} + \lambda_1 L_{align} + \lambda_2 L_{reg} \quad (8)$$

Here, $\lambda_1$ and $\lambda_2$ are weighting coefficients are used to balance the impact of various strategies on the optimization process. This joint objective ensures that the model can maintain its sensitivity to task structure with limited samples, while steadily learning the relationship between text representation and label semantics, guiding structured prompts to achieve maximum effectiveness in the latent space.

IV. EXPERIMENTAL ANALYSIS

A. Dataset

This study adopts AG News as the primary evaluation dataset. It is a representative benchmark for news-oriented text classification and is constructed from multiple authoritative news sources. The dataset covers a wide range of topics and domains, which reflects the semantic diversity present in real applications. It contains four thematic labels corresponding to world, sports, business, and technology, forming a stable and well-defined semantic structure. These characteristics make AG News suitable as a foundational test environment for few-shot text classification and prompt optimization tasks.

The text length in AG News is moderate, and the writing style is natural. It includes many short news passages composed of multiple sentences. This allows the dataset to capture the model's ability to handle mixed long and short sentence structures. Its semantic distribution is relatively balanced. The categories share some similarities while maintaining clear distinctions. This creates a challenging label semantic space for structured prompts. The dataset also provides explicit task definitions and standardized classification objectives, enabling the proposed method to be evaluated on representative natural language corpora.

Because AG News spans multiple domains, its content includes factual descriptions as well as opinion expressions and topic-related vocabulary. This setting closely simulates the multi-scenario classification demands found in practical applications. For few-shot tasks, such cross-domain semantic characteristics increase the dataset's research value within a structured prompt optimization framework. They help assess whether a model can build stable semantic understanding with limited samples and maintain consistent generalization across different label semantics.

B. Experimental Results

This paper first conducts a comparative experiment, and the experimental results are shown in Table 1.

Table 1. Comparative experimental results

| Method | Acc | Precision | Recall | AUC |
|---|---|---|---|---|
| TextConvoNet[19] | 0.842 | 0.835 | 0.828 | 0.889 |
| Transformer[20] | 0.867 | 0.861 | 0.856 | 0.913 |
| 1DCNN[21] | 0.831 | 0.824 | 0.819 | 0.874 |
| MLP[22] | 0.812 | 0.806 | 0.801 | 0.859 |
| LSTM-Transfomer[23] | 0.879 | 0.872 | 0.869 | 0.926 |
| Bert[24] | 0.894 | 0.889 | 0.883 | 0.941 |
| Ours | 0.921 | 0.915 | 0.912 | 0.964 |

Overall, traditional models show limited semantic modeling ability under few-shot settings, BERT performs better due to transferable pretrained representations but remains constrained by insufficient task guidance, while the proposed structured prompt optimization framework achieves the best and most stable performance by providing explicit semantic and category guidance, especially in terms of AUC, with learning rate sensitivity illustrated in Figure 2.

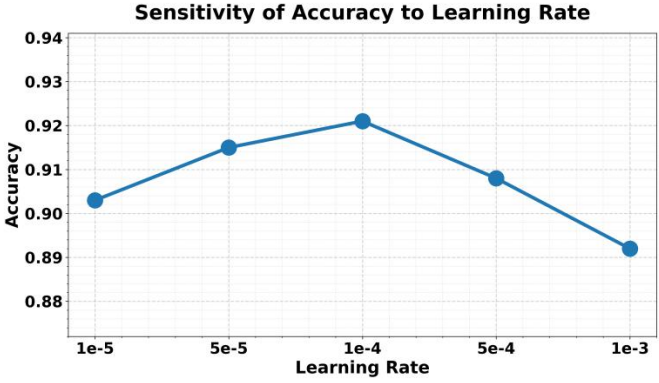

Figure 2. Hyperparameter sensitivity experiment of learning rate to the Acc metric

From the overall trend, the learning rate has a significant impact on model performance in few-shot text classification. As the learning rate increases from $1\times10^{-5}$ to $1\times10^{-4}$, the Acc value shows a stable upward trend. This indicates that within this range, the model can better use gradient information for parameter updates. It also adapts more effectively to the task semantics provided by the structured prompts, which strengthens the alignment between text and labels. This trend shows that a moderate increase in learning rate improves semantic fusion efficiency and reduces early-stage stagnation during training. When the learning rate reaches $1\times10^{-4}$, the model achieves the highest classification accuracy. This reflects that under this hyperparameter setting, the structured prompt optimization framework forms the most stable semantic separation and feature guidance in the latent space. At this point, the model maintains a balance between absorbing prompt information and modeling original text features, allowing it to establish clearer decision boundaries under few-shot conditions. This result also confirms the coupling between prompt structure and parameter update speed. A suitable learning rate maximizes the guiding ability of structured prompts. However, when the learning rate increases further to $5\times10^{-4}$, the Acc value begins to decline. This shows that an overly large learning rate damages the stability of the latent structure formed during training and weakens the alignment between text representations and label structures. The effect is more pronounced under few-shot settings, where limited data makes the model more vulnerable to gradient fluctuations. As a result, semantic shift and prompt degradation may occur. These observations also indicate that the learning rate affects not only training speed but also the stability of structured prompts across semantic subspaces.

When the learning rate reaches $1\times10^{-3}$, the performance drop becomes most evident, and the Acc value is far below the best point. This shows that the model cannot maintain fine-grained semantic control required by structured prompts under high learning rate settings. The interaction between prompt vectors and text features becomes unstable, and the latent representations fail to form clear class boundaries. Overall, the results highlight the dependence of the proposed structured prompt optimization framework on training stability. The learning rate, as a key hyperparameter, plays a decisive role in maintaining prompt effectiveness and improving few-shot classification performance.

This paper also presents an experiment on the hyperparameter sensitivity of the Prompt length to the F1-Score metric, and the experimental results are shown in Figure 3.

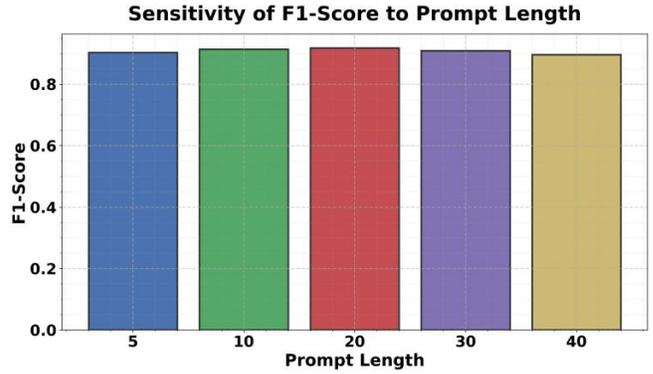

Figure 3. Hyperparameter sensitivity experiment of Prompt length to F1-Score metric

From the overall trend, the prompt length has a measurable impact on the model's F1-Score, although the variation is relatively mild. This indicates that the structured prompt framework maintains strong and stable semantic guidance under different length settings. When the prompt length increases from 5 to 20, the F1-Score shows a slight improvement. This suggests that a moderate prompt length provides richer label semantic structure, allowing the model to better align the relationship between text and category under few-shot conditions. A length of 20 achieves the highest value, showing that the structural information carried by the prompt reaches an optimal balance without being overly simplified or introducing unnecessary content.

When the prompt length continues to increase to 30 and 40, the F1-Score experiences a slight decline. This indicates that an excessively long prompt may introduce distracting semantics. Such interference can cause mild entanglement in the latent space and weaken the guiding role of structured prompts. In few-shot text classification, overly long prompts may dilute the key label features and make it harder for the model to maintain compact decision boundaries. Overall, the experiment further confirms the sensitivity of structured prompts to length design. A moderate and well-organized prompt is essential for ensuring effective semantic guidance and improving few-shot classification performance.

## V. CONCLUSION

This study addresses the common issues of semantic entanglement, weakened label structure, and insufficient feature representation in few-shot text classification. It proposes an optimization framework based on structured prompts. The framework constructs multi-dimensional prompt structures that integrate task semantics, label attributes, and textual information during the encoding stage. This makes category boundaries in the latent space more distinct. By designing systematic interactions among prompt vectors, label embeddings, and text representations, the model maintains stable semantic extraction even under extremely limited data conditions. It also improves adaptation to complex semantic patterns and cross-domain text. Overall, the proposed method offers an effective solution for few-shot text classification and lays the foundation for reducing dependence on large-scale annotated datasets. During experimentation, the structured prompt framework demonstrates strong generalization and scalability. By shifting prompts from natural language to structured forms, the model forms semantic layouts in the latent space that are more interpretable and more controllable. This helps the model maintain stable performance under style variation, domain shifts, and changes in corpus distribution. The mechanism enhances rapid adaptation to new tasks, new domains, and new label systems in practical applications. It also improves sustainability in dynamic environments. In addition, the use of structured prompts makes internal feature flows more transparent and provides a foundation for future research on interpretability enhancement, error diagnosis, and safety analysis.

Looking ahead, few-shot text classification still has broad development potential. As pretrained models continue to expand in scale and multimodal semantic structures become more integrated, the design and optimization of structured prompts will gain greater expressive flexibility. Future research may explore automatic prompt construction, cross-task prompt sharing, and coordinated optimization of multi-source structured prompts. These directions can further enhance the applicability of few-shot models in complex real-world environments. The structured prompt framework can also be extended to tasks such as question answering, reasoning, and dialogue systems, where semantic boundaries are more critical. This may provide stronger and more efficient technical support for natural language processing applications in public services, financial risk management, and medical information analysis.